\documentclass[a4paper]{article}

\usepackage{amssymb}
\usepackage{amsmath,amssymb}
\usepackage{multirow}
\usepackage{booktabs}
\usepackage{chngpage}
\usepackage{subcaption}
\usepackage{url}
\usepackage{algorithmic}
\usepackage{algorithm}
\usepackage[pdftex]{graphicx}

\title{A Self-matching Training Method with Annotation Embedding Models for Ontology Subsumption Prediction} 
\author{Yukihiro Shiraishi\footnotemark[1] \;\;\;\;\;Ken Kaneiwa\footnotemark[1]}
\date{\footnotesize	\footnotemark[1] Department of Computer and Network Engineering, Graduate School of Informatics and Engineering,
The University of Electro-Communications, Tokyo, Japan
\\
shiraishi@sw.cei.uec.ac.jp, kaneiwa@uec.ac.jp}

\begin{document}
\maketitle

\begin{abstract}
Recently, ontology embeddings representing entities in a low-dimensional space have been proposed for ontology completion. However, the ontology embeddings for concept subsumption prediction do not address the difficulties of similar and isolated entities and fail to extract the global information of annotation axioms from an ontology. In this paper, we propose a self-matching training method for the two ontology embedding models: \textbf{In}verted-index \textbf{M}atrix \textbf{E}mbedding (InME) and \textbf{Co}-occurrence \textbf{M}atrix \textbf{E}mbedding (CoME). The two embeddings capture the global and local information in annotation axioms by means of the occurring locations of each word in a set of axioms and the co-occurrences of words in each axiom. The self-matching training method increases the robustness of the concept subsumption prediction when predicted superclasses are similar to subclasses and are isolated to other entities in an ontology. Our evaluation experiments show that the self-matching training method with InME outperforms the existing ontology embeddings for the GO and FoodOn ontologies and that the method with the concatenation of CoME and OWL2Vec* outperforms them for the HeLiS ontology.
\end{abstract}

\section{Introduction}
Ontology is a data management technology that supports the sharing and reuse of formally represented knowledge \cite{Ontology}. Many ontologies have been widely used in various domains, such as readable data and semantic searches in the Semantic Web and data analysis in bioinformatics. \par
It is crucial to automatically complete ontology because of labor-intensive. Although ontology reasoners, such as HermiT \cite{HermiT} and ELK \cite{ELK}, can logically infer concept subsumption from ontologies, it is impossible to predict concept subsumption without logical connections. For link prediction, knowledge graph (KG) embeddings~\cite{TransE,DistMult,TransH,HolE,ComplEx,RotatE,HAKE,conE} have recently been studied to represent entities and relations as low-dimensional vectors. These embeddings are not effective for concept subsumption prediction because they cannot handle richer concept representations in ontologies.\par
KG embeddings have inspired the development of description logic (DL) and ontology embeddings. E2R \cite{E2R} models logical structures by applying quantum logic to $\mathcal{ALC}$. In the embeddings for description logic $\mathcal{EL}^{++}$ \cite{el}, logical structures, such as concept inclusions, are regarded as geometric operations \cite{ElEm,EmEL,BoxEL}. Onto2Vec \cite{Onto2Vec} treats logical axioms as text corpora for training Word2Vec \cite{Word2Vec1,Word2Vec2}. OPA2Vec \cite{OPA2Vec} is an embedding model that combines logical axioms and annotation axioms. OWL2Vec* \cite{OWL2Vec} converts logical axioms, annotation axioms, and RDF graph walks in an OWL ontology into text corpora. The embeddings of entities in OWL2Vec* apply to a binary classifier such as Random Forest for concept subsumption prediction. BERTSubs \cite{BERTSubs} uses the contextual word-embedding BERT \cite{BERT} for OWL ontologies instead of the non-contextual word-embedding Word2Vec. In this method, a classification layer attached to the BERT architecture is trained to predict concept subsumption. However, the previous approaches to ontology embeddings do not address the difficulties of similar and isolated entities in a binary classifier for concept subsumption prediction. In addition, the ontology embeddings capture the local information of annotation axioms but they fail to extract the global information of the axioms.\par
In this paper, we propose a self-matching training method for the two ontology embedding models: \textbf{In}verted-index \textbf{M}atrix \textbf{E}mbedding (InME) and \textbf{Co}-occurrence \textbf{M}atrix \textbf{E}mbedding (CoME). The self-matching training method enhances a binary classifier to recognize the similarity between subclasses and superclasses. As global information, InME indicates the occurring locations of each word used in a set of annotation axioms. As local information, CoME represents the co-occurrences of words in each annotation axiom. Unlike OWL2Vec* and the other models using Word2Vec and BERT, the two embeddings are simply built from only the annotation axioms, which are compressed by an autoencoder \cite{autoencoder}. We apply the self-matching training method of a binary classifier to positive and negative samples represented by the two embeddings. This method increases the robustness of the concept subsumption prediction when predicted superclasses are similar to subclasses and are isolated to other entities in an ontology.\par
In experiments, we evaluate the self-matching training method with InME and CoME for concept subsumption prediction on the OWL ontologies, GO \cite{GO}, FoodOn \cite{FoodOn}, and HeLiS \cite{Helis}. The experiments show that InME with self-matching training outperforms OWL2Vec* and other ontology embedding models for GO and FoodOn, and that CoME concatenated with OWL2Vec* outperforms the existing models for HeLiS.\par
The contributions of this study are summarized as follows.\par 
\begin{itemize}
  \item For concept subsumption prediction, the self-matching training method improves the performance of a binary classifier for predicting similar and isolated entities.
  \item The two embeddings InME and CoME can extract global and local information from annotation axioms, which characterize entities by means of the similarities and differences among general and specific words.
  \item The experiments show that InME with self-matching training outperforms the existing ontology embeddings for GO and FoodOn, while CoME concatenated with OWL2Vec* outperforms them for HeLiS.
\end{itemize}\par 
The remainder of this paper is organized as follows. In Section \ref{sec:related_works}, we describe related works on ontology embeddings. In Section \ref{sec:preliminary}, we introduce the concepts of OWL ontology, OWL2Vec*, and the autoencoder. In Section \ref{sec:preliminary_analysis}, we analyze the prediction of similar and isolated entities with a binary classifier and the contribution of logical structures and annotation axioms to concept subsumption prediction. In Section \ref{sec:propose}, we present a procedure of the self-matching training method and formalize inverted-index and co-occurrence matrices for the two embedding models InME and CoME. In Section \ref{sec:experiment}, we evaluate the performance of the self-matching training method with InME and CoME on three benchmark ontologies. Finally, we conclude the study and discuss future research in Section \ref{sec:conclusion}.

\section{Related Works \label{sec:related_works}}
\subsection{KG Embeddings}
Let $\mathbb{E}$ be a set of entities and $\mathbb{R}$ be a set of relations. A knowledge graph (KG) consists of a set of triples $(h, r, t)$, where $h, t \in \mathbb{E}$, and $r \in \mathbb{R}$. KG embeddings represent entities and relations as low-dimensional vectors for machine learning tasks such as link prediction \cite{TransE}. The KG embeddings aim to optimize these vectors to minimize a score based on the loss function for these triples. TransE \cite{TransE} is a method to interpret relations as translations operating on the low-dimensional embeddings of entities. By extending TransE, many other KG embedding models such as TransR \cite{TransR}, DistMult \cite{DistMult}, RotatE \cite{RotatE}, and HAKE \cite{HAKE} have been proposed. However, these models do not consider richer representations such as logical structures and annotation axioms.

\subsection{DL Embeddings}
Based on the KG embeddings, embedding models have been extended to description logics (DLs). E2R \cite{E2R} is defined in the interpretations of quantum logic to represent logical structures in $\mathcal{ALC}$. ELEm \cite{ElEm} and EmEL$^{++}$ \cite{EmEL} are designed for description logic $\mathcal{EL}^{++}$ by mapping logical structures to the translations of $n$-dimensional spheres. BoxEL \cite{BoxEL} uses boxes instead of spheres to improve the representation of the intersection or overlap of the $n$-dimensional geometries. However, these embedding models do not capture annotation axioms in ontologies, which lexically indicate differences and similarities among entities as classes and individuals.

\subsection{Corpus Embeddings}
\begin{table}[t]
\centering
\caption{The extracted information and embedding methods of ontology embedding models.}
\label{tab:difference_owl_model}
\resizebox{\textwidth}{!}{
\begin{tabular}{cccccc} \toprule
    \multirow{2}{*}{Model} & \multirow{2}{*}{Logical Axioms} & \multirow{2}{*}{Graph Structures} & \multicolumn{2}{c}{Annotation Axioms} & \multirow{2}{*}{Embedding Method} \\
    & & & Global Information & Local Information & \\ \midrule\addlinespace
    Onto2Vec & $\checkmark$ & $\times$ & $\times$ & $\times$ &Word2Vec \\
    OPA2Vec & $\checkmark$ & $\times$ & $\times$ & $\checkmark$ & Word2Vec \\
    OWL2Vec* & $\checkmark$ & $\checkmark$ & $\times$ & $\checkmark$ & Word2Vec \\
    BERTSubs & $\checkmark$ & $\checkmark$ & $\times$ & $\checkmark$ & BERT\\
    InME & $\times$ & $\times$ & $\checkmark$ & $\times$ & Inverted-index Matrix+Autoencoder \\
    CoME & $\times$ & $\times$ & $\times$ & $\checkmark$ & Co-occurrence Matrix+Autoencoder \\ \bottomrule
\end{tabular}
}
\end{table}
As another approach to ontology embeddings, corpus embeddings are implemented by applying word embedding models such as Word2Vec \cite{Word2Vec1,Word2Vec2} to the corpora extracted from logical and annotation axioms. Onto2Vec \cite{Onto2Vec} treats logical axioms as corpora to embed entities such as classes and individuals into low-dimensional vectors using Skip-Gram and CBOW in Word2Vec. OPA2Vec \cite{OPA2Vec} is an extended model that adds not only logical structures but also annotation axioms to the corpora. However, these two models cannot extract connections between axioms because they treat each axiom as a single sentence. OWL2Vec* \cite{OWL2Vec} attempts to resolve this problem by exploring random walks of RDF graph transformed by logical axioms in an OWL ontology. In OWL2Vec*, corpora are extracted from logical axioms, annotation axioms, and graph structures. BERTSubs \cite{BERTSubs} adopts BERT \cite{BERT} as a contextual word embedding model, instead of Word2Vec as a non-contextual model. This model fine-tunes a pre-trained language model for concept subsumption prediction. Table \ref{tab:difference_owl_model} shows the differences of the ontology embedding models whether logical axioms, graph structures, or annotation axioms are adopted. While OPA2Vec, OWL2Vec*, and CoME consider local information in annotation axioms, which is the co-occurrence of annotation words, only InME extracts global information, which is the occurring locations of words. InME and CoME are constructed from the inverted-index and co-occurrence matrices instead of Word2Vec or BERT.\par
For ontology prediction, the ontology embeddings are applied to binary classifiers and similarity measures such as Random Forest (RF), Resnik's semantic similarity measure \cite{Resnik}, Lin's semantic similarity measure \cite{Lin}, and cosine similarity to ontology prediction tasks. In concept subsumption prediction, Random Forest is reported to be optimal \cite{OWL2Vec}.\par

\section{Preliminary \label{sec:preliminary}}
\subsection{OWL Ontology}
OWL (Web Ontology Language) \cite{OWL} is a markup language for the Semantic Web to describe axioms in ontologies. OWL ontologies consist of entities such as classes, properties, and individuals, which are represented by internationalized resource identifiers\footnote{The IRIs are unique identifiers that extend the ASCII of uniform resource identifiers (URIs) to Unicode to support multiple languages.} (IRI). An OWL ontology is described by class expression axioms, object property axioms, data property axioms, and annotation axioms.\par
Let $\mathbb{C}$, $\mathbb{R}$, and $\mathbb{I}$ be sets of class names (also known as concept names), property names (also known as role names), and individual names. Class expressions are built from classes, properties, and logical constructors. A TBox $\mathbb{T}$ is a finite set of concept inclusions $C \sqsubseteq D$, where $C$ and $D$ are class expressions. An ABox $\mathbb{A}$ is a finite set of concept assertions $C(a)$ and property assertions $R(a, b)$, where $a$ and $b$ are individuals, and $R$ is a property. An OWL ontology can be represented by a finite set of RDF triples. For example, a concept inclusion axiom can be written as (\textit{obo:GO\_0021611 rdfs:subClassOf obo:GO\_0021603}). The class \textit{obo:GO\_0021611} is annotated by the property \textit{rdfs:label} and other properties such as \textit{obo:IAO\_0000115} (``definition'') as follows:
\begin{equation*}
  \begin{split}
    \textit{obo:GO\_0021611} \quad &\textit{rdfs:label} \quad \text{``facial nerve formation''}\\
    \textit{obo:GO\_0021611} \quad &\textit{obo:IAO\_0000115} \quad \text{``The process that gives rise to $\cdots$''}
  \end{split}
\end{equation*}

\subsection{OWL2Vec* \label{sec:owl2vec}}
OWL2Vec* \cite{OWL2Vec} treats graph walks, logical axioms, and annotation axioms as text corpora to learn word embeddings by Word2Vec \cite{Word2Vec1,Word2Vec2}. The graph walks are generated from an RDF graph projected from an OWL ontology. A structural document $D_{\rm{s}}$ includes logical axioms and random walks derived from the RDF graph. A lexical document $D_{\rm{l}}$ includes two kinds of word sentences. The first is the entity IRI sentences in $D_{\rm{s}}$, where each entity IRI is replaced with the annotation words of the entity. The second is the sentences extracted from annotation axioms, where words are transformed into lowercase letters, and non-English characters are removed. A combined document $D_{\rm{c}}$ is a mixture of $D_{\rm{s}}$ and $D_{\rm{l}}$. These three documents are arbitrarily combined as text corpora to train Word2Vec.\par
Let $e \in \mathbb{C} \cup \mathbb{I}$ be an entity as a class or individual in $D_{\rm{s}}$, $D_{\rm{l}}$, and $D_{\rm{c}}$ and let $w$ be a word in annotation axioms except for the IRIs. OWL2Vec* provides the two embeddings $V_{\rm{iri}}(e)$ and $V_{\rm{word}}(e)$ of each entity $e$. The first entity embedding $V_{\rm{iri}}(e)$ is defined as the word embedding $V_{\rm{w2v}}(iri(e))$ of the $e^{\prime}$s IRI, where $V_{\rm{w2v}}(w)$ is the embedding of word $w$ in Word2Vec. The second entity embedding $V_{\rm{word}}(e)$ is averaged by the word embeddings $V_{\rm{w2v}}(w)$ of words $w$ in $W_{\rm{label}}(e)$, where $W_{\rm{label}}(e)$ is the set of words in $e^{\prime}$s annotations by the property \textit{rdfs:label} $\in \mathbb{R}$.
\begin{equation}
    \label{eq:word_mean}
    V_{\rm{word}}(e) = \frac{1}{|W_{\rm{label}}(e)|} \sum_{w \in W_{\rm{label}}(e)} V_{\rm{w2v}}(w)
\end{equation}
Some IRI names are added to $W_{\rm{label}}(e)$ if an entity $e$ has no annotation word. For example, the IRI name ``Lysine'' in $W_{\rm{label}}(\textit{Lysine\_1001})$ is extracted from the IRI \textit{Lysine\_1001}\footnote{\textit{http://www.fbk.eu/ontologies/virtualcoach\#Lysine\_100}.} on HeLiS ontology \cite{OWL2Vec}.\par
In OWL2Vec*, concept subsumption is predicted by a binary classifier $f:\textbf{R}^{2\textit{n}} \rightarrow \textbf{R}$ such as Random Forest (RF), MLP, SVC, or Logistic Regression (LR), where $n$ is an embedding dimension. The following probability $p_{\rm{true}}$ indicates the likelihood of concept inclusions or concept assertions $(e_{1}, e_{2})$ being valid, where $e_{1} \in \mathbb{C} \cup \mathbb{I}$ and $e_{2} \in \mathbb{C}$.
\begin{equation}
    \label{eq:classification}
    p_{\rm{true}}(e_{1}, e_{2}) = f(V(e_{1})||V(e_{2}))
\end{equation}
where $V \in \{V_{\rm{iri}},V_{\rm{word}}\}$ and $||$ is the concatenation of two embeddings. OWL2Vec* reported that the RF is the best classifier of RF, MLP, SVC, and LR \cite{OWL2Vec}. To learn the probability $p_{\rm{true}}$, we use a training set $S_{\rm{train}} = S^{+}_{\rm{train}} \cup S^{-}_{\rm{train}}$ of pairs $(e_{1}, e_{2})$ of entities, where $S_{\rm{train}}^{+}$ is a set of positive samples and $S_{\rm{train}}^{-}$ is a set of negative samples. Each sample $(e_{1}, e_{2}) \in S_{\rm{train}}$ indicates a concept inclusion $e_{1} \sqsubseteq e_{2}$ or concept assertion $e_{2}(e_{1})$. The positive samples in $S_{\rm{train}}^{+}$ are extracted from logical axioms in an OWL ontology. For each positive sample $(e_{1}, e_{2}) \in S_{\rm{train}}^{+}$, negative samples are constructed by replacing $e_{2}$ with a randomly selected entity $e_{2}^{\prime}$ such that $(e_{1}, e^{\prime}_{2})$ $\not \in S_{\rm{train}}^{+}$, where $e_{1} \sqsubseteq e^{\prime}_{2}$ or $e^{\prime}_{2}(e_{1})$ is not logically inferred from the logical axioms. To rank each class $x \in \mathbb{C}$, the probability $p_{\rm{true}}(e_{1}, x)$ of a concept inclusion $e_{1} \sqsubseteq x$ or concept assertion $x(e_{1})$ for entity $e_{1}$ is predicted by the trained binary classifier.

\subsection{Autoencoder}
Autoencoder \cite{autoencoder} is a neural network that learns weights by minimizing the discrepancy between the original and the reconstructed data. High-dimensional data can be compressed into low-dimensional data by extracting a low-dimensional middle layer from a trained neural network. The middle layer $H \in \textbf{R}^{d \times m}$ and reconstructed output $X^{\prime} \in \textbf{R}^{n \times m}$ are given by
\begin{equation}
\label{eq:auto_input}
    H = \operatorname{ReLU}(W_{\rm{in}}X + B_{\rm{in}})
\end{equation}
\begin{equation}
\label{eq:auto_output}
    X^{\prime} = \sigma(W_{\rm{out}}H + B_{\rm{out}})
\end{equation}
where $X \in \textbf{R}^{n \times m}$ is the original input, $W_{\rm{in}} \in \textbf{R}^{d \times n}$ and $W_{\rm{out}} \in \textbf{R}^{n \times d}$ are the layer weights, $B_{\rm{in}} \in \textbf{R}^{d \times m}$ and $B_{\rm{out}} \in \textbf{R}^{n \times m}$ are the layer biases, and $\sigma$ is the sigmoid function. The loss function to minimize the discrepancy between $X$ and $X^{\prime}$ is given by
\begin{equation}
    L(X, X^{\prime}) = - \sum_{j=1}^{m} (X_{ij} \log X_{ij}^{\prime} + (1 - X_{ij}) \log (1 - X_{ij}^{\prime}))
\end{equation}

\section{Analysis of Concept Subsumption Prediction \label{sec:preliminary_analysis}}

\subsection{Prediction of Similar and Isolated Entities \label{sec:analysis_rf}}
We perform two preliminary experiments to clarify the difficulties of predicting similar and isolated entities in an OWL ontology. In the experiments, the binary classifier RF is trained on the concatenated embeddings $V_{\rm{word}}(e_{1})||V_{\rm{word}}(e_{2})$ of OWL2Vec* for concept subsumption prediction. We compare the classifier RF with distance-based ranking on the datasets GO, FoodOn, and HeLiS\footnote{Note that HeLiS is a revised version described in Section \ref{sec:datasets}}.

In the first experiment, we evaluate the trained classifier RF for predicting itself as a superclass from a randomly embedded entity $e_{r}$, i.e., $e_r \sqsubseteq e_r$ or $e_r(e_r)$. Table \ref{tab:expr_similar} shows the mean reciprocal rank (MRR) from 100 ranks by initializing the embedding of $e_{r}$ randomly 100 times. The performance of RF indicates that the classifier fails to predict superclasses similar to a subclass $e_{r}$ in all the datasets. Obviously, the distance-based ranking results in the MRR 1.0 for all the datasets because the distance of the subclass $e_{r}$ and itself as a superclass $e_{r}$ is zero.\par
The second experiment analyzes the performances for restricted test data where subclasses, superclasses, or both subclasses and superclasses are isolated entities. Let $\mathbb{E}$ be the set of all entities in $S_{\rm{train}} \cup S_{\rm{test}}$ where $S_{\rm{train}}$ is a set of training data and $S_{\rm{test}}$ is a set of test data. We extract the set of entities that appeared in the positive samples in $S_{\rm{train}}^{+}$ by $\mathbb{E}^{+} = \{ e_{1}, e_{2} \in{\mathbb{E}}|(e_{1}, e_{2}) \in S_{\rm{train}}^{+} \}$. An entity $e$ is an isolated entity if $e$ does not occur in $S_{\rm{train}}^{+}$, i.e., $e \not\in \mathbb{E}^{+}$. We denote $\mathbb{E}^{I}$ as the set of isolated entities $e \in \mathbb{E} \setminus \mathbb{E}^{+}$. Some of the isolated entities not in the positive samples may be contained in negative samples because the entities $e^{\prime}_{2}$ of negative samples $(e_{1}, e^{\prime}_{2})$ are randomly selected. Table \ref{tab:expr_split_test} shows the MRRs of the classifier RF and distance-based ranking for each restricted test data on $\mathbb{E}^{+} \!\!\times\! \mathbb{E}^{+}$, $\mathbb{E}^{I} \!\times\! \mathbb{E}^{+}$, $\mathbb{E}^{+} \!\!\times\! \mathbb{E}^{I}$, and $\mathbb{E}^{I} \!\times\! \mathbb{E}^{I}$. The RF outperforms the distance-based ranking when entities in $\mathbb{E}^{+}$ are predicted as superclasses for $S_{\rm{test}}$ on $\mathbb{E}^{I} \!\times\! \mathbb{E}^{+}$. However, the RF significantly underperforms the distance-based ranking for $S_{\rm{test}}$ on $\mathbb{E}^{+} \!\times\! \mathbb{E}^{I}$ and $\mathbb{E}^{I} \!\times\! \mathbb{E}^{I}$ because the predicted superclasses are isolated entities in $\mathbb{E}^{I}$. Thus, the RF poorly predicts isolated entities but the distance-based ranking is suitable for the prediction. From these results, it is required to take advantage of both RF and distance-based ranking.
\begin{table}[t]
  \caption{The performance of the classifier RF and distance-based ranking with OWL2Vec* on the self-prediction.}
  \label{tab:expr_similar}
  \begin{center}
  \resizebox{8cm}{!}{
    \begin{tabular}{cc cc cc} \toprule
    \multicolumn{2}{l}{GO} & \multicolumn{2}{l}{FoodOn} & \multicolumn{2}{l}{HeLiS} \\
    \cmidrule(lr){1-2} \cmidrule(lr){3-4} \cmidrule(lr){5-6}
    RF & Distance & RF & Distance & RF & Distance \\
    MRR & MRR & MRR & MRR & MRR & MRR \\ \midrule
    0.002 & \bf{1.0} & 0.002 & \bf{1.0} & 0.012 & \bf{1.0} \\ \bottomrule
    \end{tabular}
    }
\end{center}
\end{table}
\begin{table}[htb]
\caption{The performance of the classifier RF and distance-based ranking with OWL2Vec* on restricted test data.}
\label{tab:expr_split_test}
\begin{center}
\resizebox{\textwidth}{!}{
\begin{tabular}{l ccc ccc ccc } \toprule
     & \multicolumn{3}{l}{GO} & \multicolumn{3}{l}{FoodOn} & \multicolumn{3}{l}{HeLiS} \\
    \cmidrule(lr){2-4} \cmidrule(lr){5-7} \cmidrule(lr){8-10}
     &  & RF & Distance &  & RF & Distance &  & RF & Distance \\
    Test Data & Num of Data & MRR & MRR & Num of Data & MRR & MRR & Num of Data & MRR & MRR \\ \midrule
    $S_{\rm{test}}$ & 14521 & 0.150 & \bf{0.157} & 5957 & \bf{0.198} & 0.125 & 435 & \bf{0.583} & 0.216 \\
    $S_{\rm{test}}$ on $\mathbb{E}^{+} \!\!\times\! \mathbb{E}^{+}$ & 9702 & 0.136 & \bf{0.160} & 1543 & \bf{0.158} & 0.102 & 0 & - & - \\
    $S_{\rm{test}}$ on $\mathbb{E}^{I} \!\!\times\! \mathbb{E}^{+}$ & 4456 & \bf{0.187} & 0.135 & 4235 & \bf{0.220} & 0.126 & 429 & \bf{0.590} & 0.215 \\
    $S_{\rm{test}}$ on $\mathbb{E}^{+} \!\!\times\! \mathbb{E}^{I}$ & 203 & 0.041 & \bf{0.351} & 59 & 0.013 & \bf{0.173} & 0 & - & - \\
    $S_{\rm{test}}$ on $\mathbb{E}^{I} \!\!\times\! \mathbb{E}^{I}$ & 160 & 0.043 & \bf{0.355} & 120 & 0.003 & \bf{0.341} & 6 & 0.030 & \bf{0.285} \\
    \bottomrule
\end{tabular}
}
\end{center}
\end{table}

\subsection{Logical Structures vs Annotations}
For concept subsumption prediction, ontology embedding models exploit the corpora extracted from logical structures and annotation axioms in an OWL ontology, as shown in Table \ref{tab:difference_owl_model}. To analyze the effect of entity embeddings, we compare the logical axioms and annotation axioms used to learn word embeddings from their extracted corpora. \par
\begin{figure}
  \centering
  \includegraphics[width=10cm]{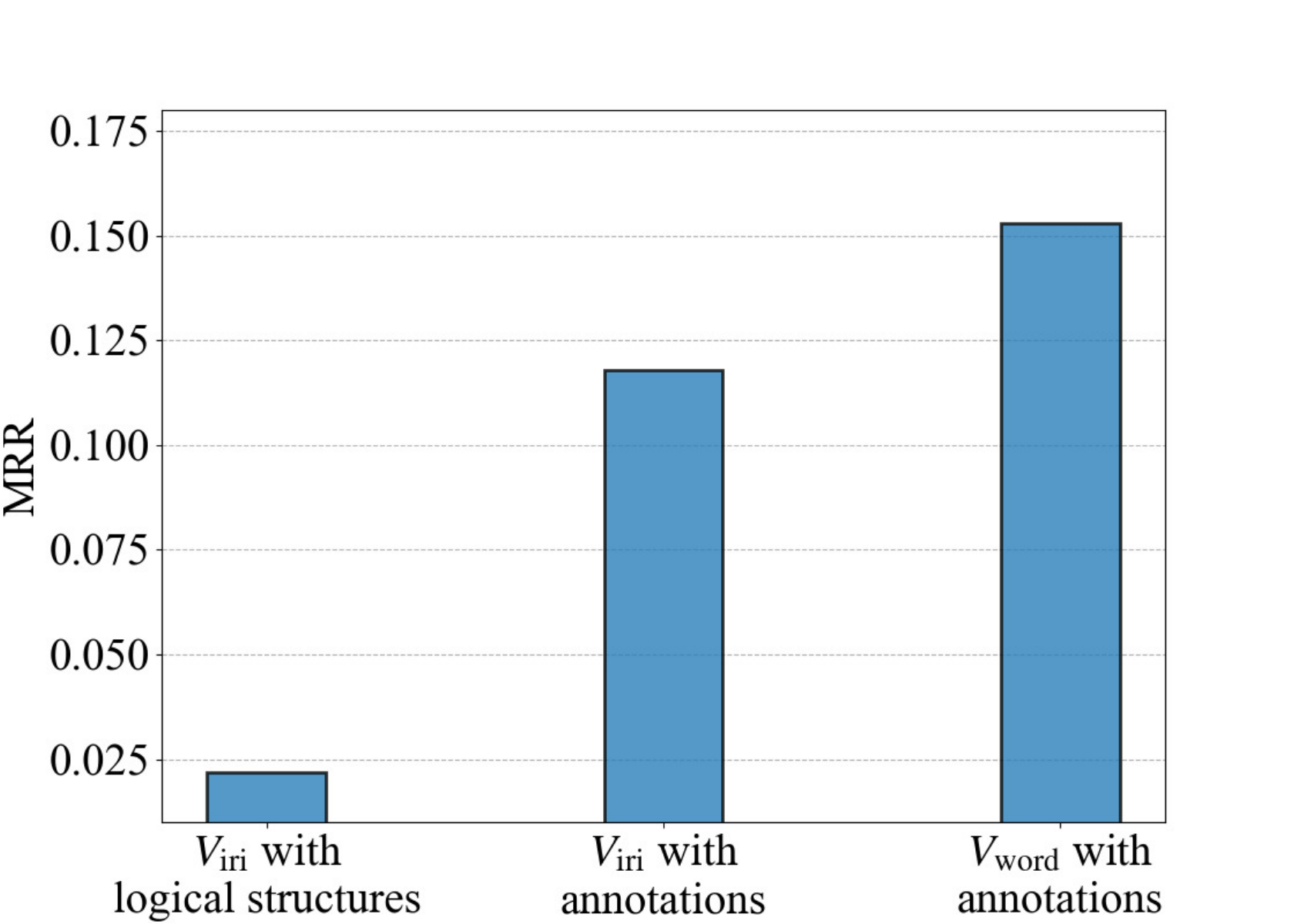}
  \caption{The MRRs of Word2Vec for logical structures and annotations on GO.}
  \label{fig:compare_intro}
\end{figure}
Figure \ref{fig:compare_intro} shows the performances of the entity embeddings $V_{\rm{iri}}$ and $V_{\rm{word}}$ with logical axioms and annotation axioms. We do not apply the entity embedding $V_{\rm{word}}$ to the logical axioms because $W_{\rm{label}}(e)=\emptyset$, i.e., no annotation words with \textit{rdfs:label}. The word embeddings are learned from the corpora of logical axioms or annotation axioms using Word2Vec. A binary classifier for concept subsumption prediction is trained on the concatenation $V(e_{1}) \Vert V(e_{2})$ of the entity embeddings $V \in \{ V_{\rm{iri}}, V_{\rm{word}} \}$. In Figure \ref{fig:compare_intro}, the mean reciprocal rank (MRR) for annotation axioms greatly outperforms that for logical axioms. Surprisingly, the corpora of logical axioms seem ineffective for the prediction. In addition, the MRR of $V_{\rm{word}}$ with annotations outperforms $V_{\rm{iri}}$ with annotations. These results suggest that ($\mathrm{i}$) only the corpora of annotation axioms contribute to concept subsumption prediction and ($\mathrm{ii}$) the entity embedding $V_{\rm{word}}$ improves the performance. 

\begin{figure*}
  \includegraphics[width=\textwidth]{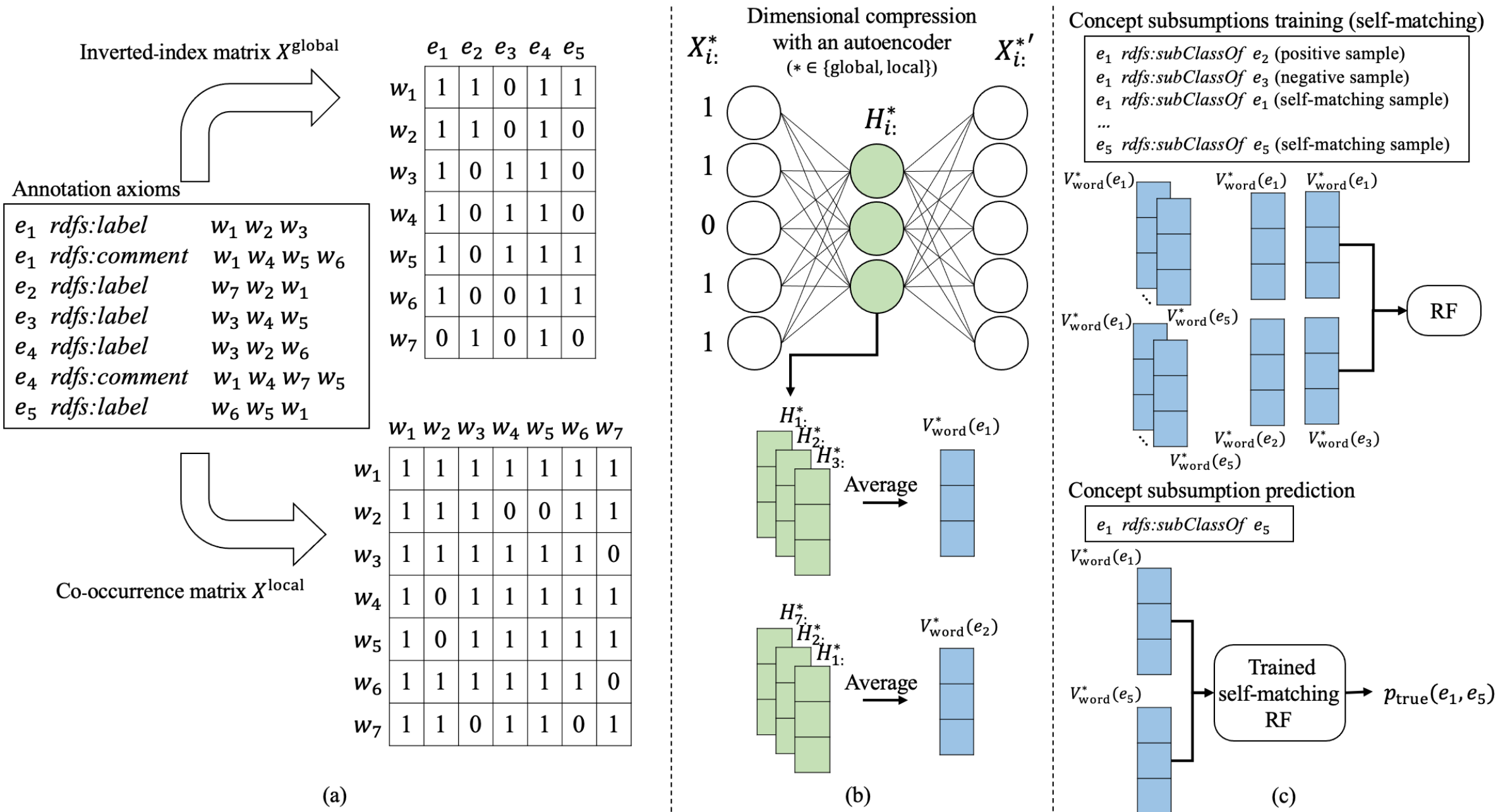}
  \caption{The framework of a self-matching training method with InME and CoME.
  (a) The inverted-index and co-occurrence matrices, $X^{\rm{global}}$ and $X^{\rm{local}}$, are constructed from annotation axioms.
  (b) The word embeddings of $H_{i}^{\ast}$ are obtained from the matrices compressed with an autoencoder and entity embeddings $V_{\rm{word}}^{\ast}$ are transformed by averaging the word embeddings.
  (c) An RF classifier is trained by the self-matching training method with InME or CoME.} \label{fig_architecture}
\end{figure*}

\section{Methodology\label{sec:propose}}
Figure \ref{fig_architecture} shows the framework of a self-matching training method with the two embeddings InME and CoME, which is applied to an RF classifier for concept subsumption prediction.

\subsection{Self-matching Training}
\begin{figure}[t]
  \begin{center}
    \includegraphics[clip,width=\textwidth]{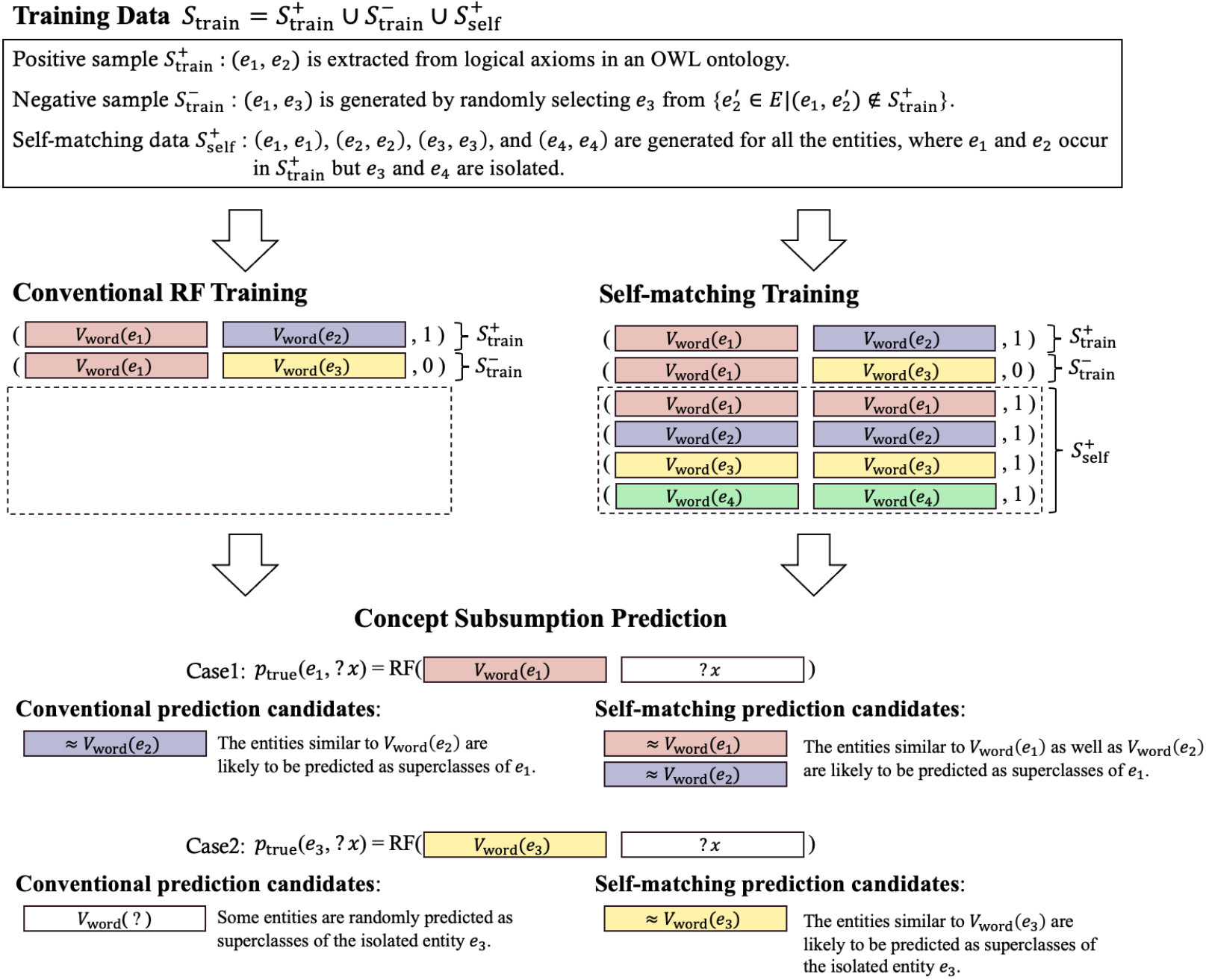}
      \caption{The flowchart of the conventional RF training and self-matching training.}
      \label{fig:self-matching}
  \end{center}
\end{figure}

We describe a flowchart of the self-matching training method for concept subsumption prediction in Figure \ref{fig:self-matching}. First of all, we obtain the set $S_{\rm{train}}^{+}$ of positive samples $(e_{1}, e_{2})$ from logical axioms in an OWL ontology. For each positive sample $(e_{1}, e_{2}) \in S_{\rm{train}}^{+}$, we add a negative sample $(e_{1}, e_{3})$ to $S_{\rm{train}}^{-}$, where $e_{3}$ is randomly selected from $\{ e_{2}^{\prime} \in \mathbb{E} | (e_{1}, e_{2}^{\prime}) \not \in S_{\rm{train}}^{+} \}$. In addition to the positive and negative samples, the set $S_{\rm{self}}^{+} = \{ (e_{i}, e_{i})|e_{i} \in \mathbb{C} \cup \mathbb{I} \}$ of new self-matching samples is generated from all the classes and individuals in the ontology, where $|S_{\rm{self}}^{+}| = |\mathbb{C} \cup \mathbb{I}|$. To recognize the similarity between subclasses and superclasses, the concatenated embeddings $V_{\rm{word}}(e_{i}) \Vert V_{\rm{word}}(e_{i})$ of each self-matching sample $(e_{i}, e_{i})$ in $S_{\rm{self}}^{+}$ are supplemented as positive samples. The RF classifier is trained on the extended training data $S_{\rm{train}} = S_{\rm{train}}^{+} \cup S_{\rm{self}}^{+} \cup S_{\rm{train}}^{-}$.\par
In Figure \ref{fig:self-matching}, entities $e_{1}$ and $e_{2}$ occur in $S_{\rm{train}}^{+}$ but entities $e_{3}$ and $e_{4}$ are isolated in the ontology, i.e., $e_{3}, e_{4} \in \mathbb{E}^{I}$ are not used in the original training set $S_{\rm{train}}^{+}$. As conventional prediction candidates of the probability $p_{\rm{true}}(e_{1}, ?x)$, the entities similar to $V_{\rm{word}}(e_{2})$ are likely to be predicted as superclasses of $e_{1}$ because of the positive sample $(e_{1}, e_{2})$. However, for the probability $p_{\rm{true}}(e_{3}, ?x)$, some entities are randomly predicted as superclasses of the isolated entity $e_{3}$ because the relationships to any isolated entity are not trained in positive samples. As shown in Section \ref{sec:analysis_rf}, the conventional RF training struggles to predict isolated entities as superclasses for the restricted test data on $\mathbb{E}^{+} \times \mathbb{E}^{I}$ and $\mathbb{E}^{I} \times \mathbb{E}^{I}$.
On the other hand, the self-matching training increases the probability $p_{\rm{true}}(e_{i}, e_{i})$ of concept inclusions and concept assertions for all the entities $e_{i}$ including isolated entities. As self-matching prediction candidates of the probability $p_{\rm{true}}(e_{1}, ?x)$, the entities similar to $V_{\rm{word}}(e_{1})$ as well as $V_{\rm{word}}(e_{2})$ are likely to be predicted as superclasses of $e_{1}$ due to the positive sample $(e_{1}, e_{2})$ and self-matching sample $(e_{1}, e_{1})$. For the probability $p_{\rm{true}}(e_{3}, ?x)$, the entities similar to $V_{\rm{word}}(e_{3})$ are likely to be predicted as superclasses of the isolated entity $e_{3}$ because of the self-matching sample $(e_{3}, e_{3})$.\par
The self-matching training method is more effective when the embeddings of a subclass and a superclass are similar to each other. For the method, we will provide two entity embeddings InME and CoME that capture the similarity between a subclass and a superclass from the global and local information of annotation axioms.

\subsection{Global and Local Information in Annotations}
We consider global and local information in annotation axioms of the following Gene Ontology \cite{GO}:
\begin{equation*}
  \begin{split}
    \textit{obo:GO\_0021603} \quad &\textit{rdfs:label} \quad \text{``cranial} \: \text{nerve} \: \text{formation''}\\
    \textit{obo:GO\_0021611} \quad &\textit{rdfs:label} \quad \text{``facial} \: \text{nerve} \: \text{formation''}\\
    \textit{obo:GO\_0021620} \quad &\textit{rdfs:label} \quad \text{``hypoglossal} \: \text{nerve} \: \text{formation''}
  \end{split}
\end{equation*}
For the ontology, the common superclass \textit{obo:GO\_0021603} has the two subclasses \textit{obo:GO\_0021611} and \textit{obo:GO\_0021620} as follows:
\begin{equation*}
  \begin{split}
  \textit{obo:GO\_0021611} \quad &\sqsubseteq \quad \textit{obo:GO\_0021603}\\
  \textit{obo:GO\_0021620} \quad &\sqsubseteq \quad \textit{obo:GO\_0021603}
  \end{split}
\end{equation*}
The annotation axioms of superclasses and subclasses frequently exhibit resemblances. Conversely, the difference and similarity among annotation axioms provide highly valuable information for predicting their relationships, regardless of the absence of logical connections. To capture the similarity among annotation axioms, each entity embedding is calculated by averaging the word embeddings (Equation (\ref{eq:word_mean})). For example, \textit{obo:GO\_0021603}'s embedding is generated by averaging the word embeddings of ``cranial,'' ``nerve,'' and ``formation.'' Similarly, \textit{obo:GO\_0021611} is the average of ``facial,'' ``nerve,'' and ``formation,'' and \textit{obo:GO\_0021620} is the average of ``hypoglossal,'' ``nerve,'' and ``formation.'' These entity embeddings tend to be similar due to the shared words ``nerve'' and ``formation.'' However, if the word embeddings of the non-shared words like ``cranial,'' ``facial,'' and ``hypoglossal'' are widely separated in the embedding space, the averaging process might prevent entity embeddings from getting close to each other. Therefore, it is important that the word embeddings are distributed close together.\par
We present to characterize the words ``cranial,'' ``facial,'' and ``nerve'' from the annotation axioms.
As local information, these words co-occur as follows:\footnote{Each of the words ``cranial'', ``facial'', and ``nerve'' is itself included in the co-occurrence words to make the features of each word more similar.}
\begin{equation*}
  \begin{split}
    \text{``cranial'' co-occurs with} &: \text{cranial, nerve, formation}\\
    \text{``facial'' co-occurs with} &: \text{facial, nerve, formation}\\
    \text{``nerve'' co-occurs with} &: \text{cranial, facial, hypoglossal,}\\
    &  \;\;\; \text{nerve, formation}
  \end{split}
\end{equation*}
These words share similarities in terms of co-occurrence with the same words.
In this example, ``cranial'' co-occurs with three words, matching the number of words in the annotation axiom of \textit{obo:GO\_0021603}. The word ``cranial'' might also appear in different annotation axioms. Hence, local information frequently contains a variety of lexical information, which can also be extracted by Word2Vec with a maximum window size.\par
Each annotation axiom contains several words to explain an entity in an ontology. As global information, the locations of the words ``cranial,'' ``facial,'' and ``nerve'' are identified by entities in a set of annotation axioms as follows:
\begin{equation*}
  \begin{split}
    \text{``cranial'' appears in} &: \textit{obo:GO\_0021603}\\
    \text{``facial'' appears in} &: \textit{obo:GO\_0021611}\\
    \text{``nerve'' appears in} &: \textit{obo:GO\_0021603, obo:GO\_0021611, obo:GO\_0021620}
  \end{split}
\end{equation*}
Similar global information is shared by words in the annotation axiom of each entity. In this example, the number of locations of ``cranial'' appearing in $\textit{obo:GO\_0021603}$'s annotation axiom is fewer compared to the number of co-occurrences of ``cranial'' as local information. In addition, the number of locations of the general word ``nerve'' is more than the specific word ``cranial.'' The local information may reduce the similarity among words because they contain excessive details due to co-occurring words. In contrast, the global information represents the difference between specific and general words. The similarity between specific and general words is less likely to decrease and the similarity between specific words is less likely to increase. Therefore, the global information characterizes more concise and essential similarities of words. As Word2Vec primarily captures local information, it is unable to extract global information, which pertains to pure entity-to-word relationships. 

\subsection{Inverted-index and Co-occurrence Matrices}
InME extracts global information between entities $e_{j} \in \mathbb{C} \cup \mathbb{I}$ and words $w_{i} \in W$, where $W$ is the set of words in the annotation axioms of entities $e_{j}$ excluding IRIs. An inverted-index matrix is defined as $X^{\rm{global}} \in \textbf{R}^{|W| \times |\mathbb{C} \cup \mathbb{I}|}$ (in Figure \ref{fig_architecture}(a)), where each $X_{ij}^{\rm{global}}$ indicates whether the word $w_{i}$ appears in the annotations of entity $e_{j}$. Let $W_{\rm{ann}}(e)$ be the set of words used in all annotation axioms of an entity $e \in \mathbb{C} \cup \mathbb{I}$, where each annotation property is not limited to \textit{rdfs:labels}. Then, $X_{ij}^{\rm{global}}$ is given by\par
\begin{equation}
\label{eq:global}
X_{ij}^{\rm{global}} =
\begin{cases}
  1 & \text{if}\ w_{i} \in W_{\rm{ann}}(e_{j}) \\
  0 & \text{otherwise}
\end{cases}
\end{equation}\par
CoME extracts local information about which words $w_{i}$ and $w_{j}$ appear together in an annotation axiom. A co-occurrence matrix is defined as $X^{\rm{local}} \in \textbf{R}^{|W| \times |W|}$ (in Figure \ref{fig_architecture}(a)), where each $X_{ij}^{\rm{local}}$ indicates the co-occurrence of word $w_{i}$ with word $w_{j}$. Then, $X_{ij}^{\rm{local}}$ is given by\par
\begin{equation}
    \label{eq:local}
    X_{ij}^{\rm{local}} = 
    \begin{cases}
     1 & \text{if}\ \exists e \in \mathbb{C} \cup \mathbb{I}.\{w_{i}, w_{j}\} \subseteq W_{\rm{ann}}(e)\\
     0 & \text{otherwise}
    \end{cases}
\end{equation}\par
The dimensions $|\mathbb{C} \cup \mathbb{I}|$ and $|W|$ of $X_{i}^{\rm{global}}$ and $X_{i}^{\rm{local}}$ are too large to be applied to a binary classifier, where $X_{i}^{\ast}$ is the $i$-th row vector of $X^{\ast}$. We transform the matrices $X^{\rm{global}}$ and $X^{\rm{local}}$ into low-dimensional word embeddings by the autoencoder (in Figure \ref{fig_architecture}(b)). The low-dimensional middle layer $H^{\ast}$ is given by
\begin{equation}
    H^{\ast} = \operatorname{ReLU}(\hat{W}_{\rm{in}} X^{\ast} + \hat{b}_{\rm{in}})
\end{equation}
where $\ast \in \{\rm{global}, \rm{local} \}$.

\subsection{Entity Embeddings and Concatenation}
Similar to OWL2Vec*, we convert word embeddings into entity embeddings by averaging the row vectors $H^{\ast}_{i:}$ for all $w_{i} \in W_{\rm{label}}(e)$. The embedding $V_{\rm{word}}^{\ast}$ of entity $e$ is defined as
\begin{equation}
    \label{eq:word_mean_our}
    V^{\ast}_{\rm{word}}(e) = \frac{1}{|W_{\rm{label}}(e)|} \sum_{w_{i} \in W_{\rm{label}}(e)} H^{\ast}_{i:}
\end{equation}

For each pair $(e_{1}, e_{2})$ of entities in $S_{\rm{train}}$, a binary classifier RF is trained by the concatenation of entity embeddings $V_{\rm{word}}^{\ast}(e_{1})$ and $V_{\rm{word}}^{\ast}(e_{2})$ (in Figure \ref{fig_architecture}(c)). The probability $p_{\rm{true}}(e_{1}, e_{2})$ of a concept inclusion $e_{1} \sqsubseteq e_{2}$ and concept assertion $e_{2}(e_{1})$ is given by
\begin{equation}
    p_{\rm{true}}(e_{1}, e_{2}) = \operatorname{RF}(V^{\ast}_{\rm{word}}(e_{1}) || V^{\ast}_{\rm{word}}(e_{2}))
\end{equation}

As the concatenation of InME and CoME, both global and local information can be considered as $V^{\rm{global}}_{\rm{word}}(e) || V^{\rm{local}}_{\rm{word}}(e)$ for each entity $e$.

\section{Experiments\label{sec:experiment}}
We evaluate our self-matching training method with InME and CoME for concept subsumption prediction on OWL ontologies.

\subsection{Datasets \label{sec:datasets}}
We use the OWL ontologies, GO \cite{GO}, FoodOn \cite{FoodOn}, and HeLiS \cite{Helis}, summarized in Table \ref{tab:data}. GO represents the biological knowledge of genes and their products expressed by DL $\mathcal{SRI}$. FoodOn represents the knowledge of foods that contain materials consumed by humans and domesticated animals, as expressed by DL $\mathcal{SRIQ}$. HeLiS represents the knowledge about food and physical activity domains, expressed by DL $\mathcal{ALCHIQ(D)}$. Whereas HeLiS contains both concept inclusion and concept assertion axioms, GO and FoodOn contain only concept inclusion axioms. We use the datasets used in OWL2Vec*\footnote{\url{https://github.com/KRR-Oxford/OWL2Vec-Star/tree/master/case_studies}}.\par
\begin{table*}[t]
\caption{The statistics of GO, FoodOn, and HeLiS, where ``Words per entity'' denotes the average number of annotation words per entity after preprocessing. The parenthesis number on HeLiS represents the case where IRI names are treated as annotation words.}
\label{tab:data}
\begin{center}
\resizebox{\textwidth}{!}{
\begin{tabular}{l rrrrrrr} \toprule
    Ontology & Classes & Individuals & Annotation axioms & Words per entity & Concept inclusions & Concept assertions \\ \midrule
    GO & 44,244 & 0 & 452,028 & 43.54 & 72,601 & 0 \\
    FoodOn & 28,182 & 0 & 142,536 & 30.07 & 29,778 & 0 \\
    HeLiS & 277 & 20,318 & 4,984 & 0.40(1.83) & 261 & 20,318 \\ \bottomrule
\end{tabular}
}
\end{center}
\end{table*}
Notice that OWL2Vec* extracts some names from the IRIs of entities that are added to the annotation words in HeLiS as described in Section \ref{sec:owl2vec}. This extraction process causes some entities obviously ranked first in testing. For example, for the concept assertion axiom $Lysine(Lysine\_100)$, the name ``Lysine'' is added as each annotation word of the entities $Lysine\_100$ and $Lysine$ because these entities have no annotation axiom; the embeddings of $Lysine\_100$ and $Lysine$ are identical because of averaging the one-word embedding of ``Lysine''. Therefore, we exclude the trivial concept assertions from validation and test data on HeLiS.

\subsection{Experimental Setup}
For the evaluation of the self-matching training method, we adopt the entity embeddings InME and CoME generated only from the annotation axioms of each ontology. Note that all words in the annotation axioms are transformed into lowercase letters, and non-English characters are removed. Concept inclusion axioms in GO and FoodOn and concept assertion axioms in HeLiS are randomly divided into training (70\%), validation (10\%), and test (20\%). We fine-tune the hyper-parameters on the validation data. For each pair $(e_{1}, e_{2})$ of entities in the validation and test data, every class $x \in \mathbb{C}$ is sorted in descending order based on the probability $p_{\rm{true}}(e_{1}, x)$, where class $x$ is filtered out from the ranking if it has already appeared in the positive training data including the self-matching data. We report the mean reciprocal rank (MRR) and Hits@$n$ $(n = 1, 5, 10)$ to evaluate each embedding model and the combinations InME $||$ CoME, InME $||$ $X$, and CoME $||$ $X$ of two embeddings for each $X \in \{$ Word2Vec, OWL2Vec* $\}$, where $||$ is the concatenation of two embeddings and Word2Vec is trained with Skip-Gram by annotation axioms. We choose $D_{\rm{s,l}}$ for GO, $D_{\rm{s,l}}$ for FoodOn, and $D_{\rm{s,l,rc}}$ for HeLiS as the best combinations of training corpora for OWL2Vec* \cite{OWL2Vec}.\par
We set the best-compressed dimension $n$ of InME and CoME for the autoencoder among $n = \{$50, 100, 200$\}$ by the highest MRR on the validation data. The self-matching training is also determined to apply by the performance on the validation data.\par
We compare our method with the KG embedding models RDF2Vec \cite{RDF2Vec}, TransE \cite{TransE}, TransR \cite{TransR}, DistMult \cite{DistMult}, the description logic $\mathcal{EL}^{++}$ embedding model ELEm \cite{ElEm}, and the ontology embedding models Onto2Vec \cite{Onto2Vec}, OPA2Vec \cite{OPA2Vec}, and OWL2Vec* \cite{OWL2Vec}. We exclude BERTSubs \cite{BERTSubs} from the comparison because it is only pre-trained for additional large texts. For the KG embedding models, OWL ontologies are converted to RDF triples by the OWL Mapping to RDF Graphs\footnote{\url{https://www.w3.org/TR/owl2-mapping-to-rdf/}}, where blank nodes are regarded as new entities. To train ELEm, we extract the description logic $\mathcal{EL}^{++}$ axioms of $C \sqsubseteq D$, $C \sqcap D \sqsubseteq E$, $C \sqsubseteq \exists R.D$, $\exists R.C \sqsubseteq D$, and $C \sqcap D \sqsubseteq \bot$ converted from OWL ontologies. The embeddings provided by each model are applied to the RF classifier for concept subsumption prediction. Note that each predicted entity is ranked in all the classes, where blank nodes, annotation words, and the positive training data including the self-matching data are filtered. We employ the implementation of OpenKE \cite{OpenKE} for TransE, TransR, and DistMult. For ELEm and OWL2Vec*, we use the source codes cited in the papers \cite{ElEm,OWL2Vec}. For RDF2Vec, Onto2Vec, and OPA2Vec, we use the implementation of OWL2Vec*\footnote{\url{https://github.com/KRR-Oxford/OWL2Vec-Star/tree/master}}.

\subsection{Results \label{sec:result}}
\begin{table}[t]
    \centering
    \caption{The results of ontology embedding models on GO, FoodOn, and HeLiS.}
    \label{tab:result_compare}
    \begin{adjustwidth}{-1in}{-1in}
    \begin{center}
    \resizebox{1.0\textwidth}{!}{      
    \begin{tabular}{lccccc ccccc cccc} \toprule
        & \multicolumn{5}{l}{GO} & \multicolumn{5}{l}{FoodOn} & \multicolumn{4}{l}{HeLiS} \\
        \cmidrule(lr){2-5} \cmidrule(lr){7-10} \cmidrule(lr){12-15}
        Embedding Model & MRR & Hits@1 & Hits@5 & Hits@10 & \hspace{10pt} & MRR & Hits@1 & Hits@5 & Hits@10 & \hspace{10pt} & MRR & Hits@1 & Hits@5 & Hits@10\\ \midrule
        RDF2Vec\cite{RDF2Vec} & 0.061 & 0.031 & 0.082 & 0.118 &&         0.076 & 0.050 & 0.092 & 0.121 && 0.384 & 0.255 & 0.501 & 0.625 \\
        TransE\cite{TransE} & 0.069 & 0.034 & 0.095 & 0.137 &&       0.103 & 0.071 & 0.128 & 0.158 && 0.398 & 0.276 & 0.529 & 0.701 \\
        TransR\cite{TransR} & 0.037 & 0.013 & 0.053 & 0.083 &&       0.070 & 0.047 & 0.081 & 0.109 && 0.384 & 0.267 & 0.485 & 0.653 \\
        DistMult\cite{DistMult} & 0.064 & 0.033 & 0.087 & 0.120 &&     0.112 & 0.077 & 0.139 & 0.180 && 0.328 & 0.214 & 0.423 & 0.618 \\
        ELEm\cite{ElEm} & 0.023 & 0.009 & 0.029 & 0.044 &&         0.066 & 0.035 & 0.090 & 0.130 && 0.113 & 0.032 & 0.177 & 0.290 \\
        Onto2Vec\cite{Onto2Vec} & 0.024 & 0.009 & 0.032 & 0.051 &&     0.048 & 0.023 & 0.069 & 0.093 && 0.333 & 0.166 & 0.478 & 0.660 \\
        OPA2Vec\cite{OPA2Vec} & 0.112 & 0.058 & 0.155 & 0.218 &&      0.149 & 0.097 & 0.193 & 0.249 && 0.351 & 0.211 & 0.487 & 0.715\\
        OWL2Vec*\cite{OWL2Vec} & 0.150 & 0.065 & 0.222 & 0.321 &&     0.198 & 0.127 & 0.271 & 0.343 && 0.583 & 0.430 & 0.777 & \underline{\bf{0.880}} \\ \midrule
        InME & \underline{\bf{0.253}} & \underline{\bf{0.148}} & \underline{\bf{0.362}} & \underline{\bf{0.478}} && \bf{0.282} & \bf{0.183} & \bf{0.394} & \bf{0.474} && 0.570 & 0.425 & 0.756 & 0.851 \\
        CoME & \bf{0.193} & \bf{0.084} & \bf{0.309} & \bf{0.421} && \bf{0.229} & \bf{0.147} & \bf{0.323} & \bf{0.389} && 0.549 & 0.418 & 0.697 & 0.779 \\
        InME$\Vert$CoME & \bf{0.240} & \bf{0.129} & \bf{0.359} & \bf{0.474} && \bf{0.292} & \bf{0.198} & \bf{0.399} & \bf{0.469} && 0.566 & \bf{0.434} & 0.740 & 0.830 \\
        InME$\Vert$Word2Vec & \bf{0.225} & \bf{0.110} & \bf{0.351} & \bf{0.476} && \underline{\bf{0.306}} & \underline{\bf{0.205}} & \underline{\bf{0.419}} & \underline{\bf{0.501}} && 0.540 & 0.402 & 0.680 & 0.811 \\
        CoME$\Vert$Word2Vec & \bf{0.198} & \bf{0.083} & \bf{0.323} & \bf{0.449} && \bf{0.274} & \bf{0.180} & \bf{0.381} & \bf{0.457} && 0.517 & 0.386 & 0.669 & 0.782 \\
        InME$\Vert$OWL2Vec* & \bf{0.225} & \bf{0.112} & \bf{0.344} & \bf{0.468} && \bf{0.282} & \bf{0.190} & \bf{0.385} & \bf{0.468} && \bf{0.601} & \bf{0.462} & 0.759 & 0.862 \\
        CoME$\Vert$OWL2Vec* & \bf{0.194} & \bf{0.086} & \bf{0.305} & \bf{0.417} && \bf{0.248} & \bf{0.161} & \bf{0.340} & \bf{0.417} && \underline{\bf{0.623}} & \underline{\bf{0.492}} & \underline{\bf{0.782}} & 0.862 \\ \bottomrule
    \end{tabular}
    }
\end{center}
\end{adjustwidth}
\end{table}
Table \ref{tab:result_compare} shows the performance of our method compared with the baselines of ontology embeddings. For GO and FoodOn, any combination of our embeddings outperforms the baselines. In particular, InME significantly outperforms the best baseline model OWL2Vec* despite only extracting the global information of annotation axioms. This implies that the global information plays a more important role in the prediction than the logical and graph structures used in OWL2Vec*. However, InME underperforms OWL2Vec* for the revised HeLiS, because HeLiS originally has extremely few annotation axioms compared to GO and FoodOn, as shown in Table \ref{tab:data}. Instead, the concatenation of CoME and OWL2Vec* outperforms OWL2Vec* in MRR and Hits@1. This result shows that our models can enhance OWL2Vec* by the fact that the co-occurrences of words in CoME supplement the lack of annotations in HeLiS.\par

%
\begin{figure*}
  \centering
  \includegraphics[width=\textwidth]{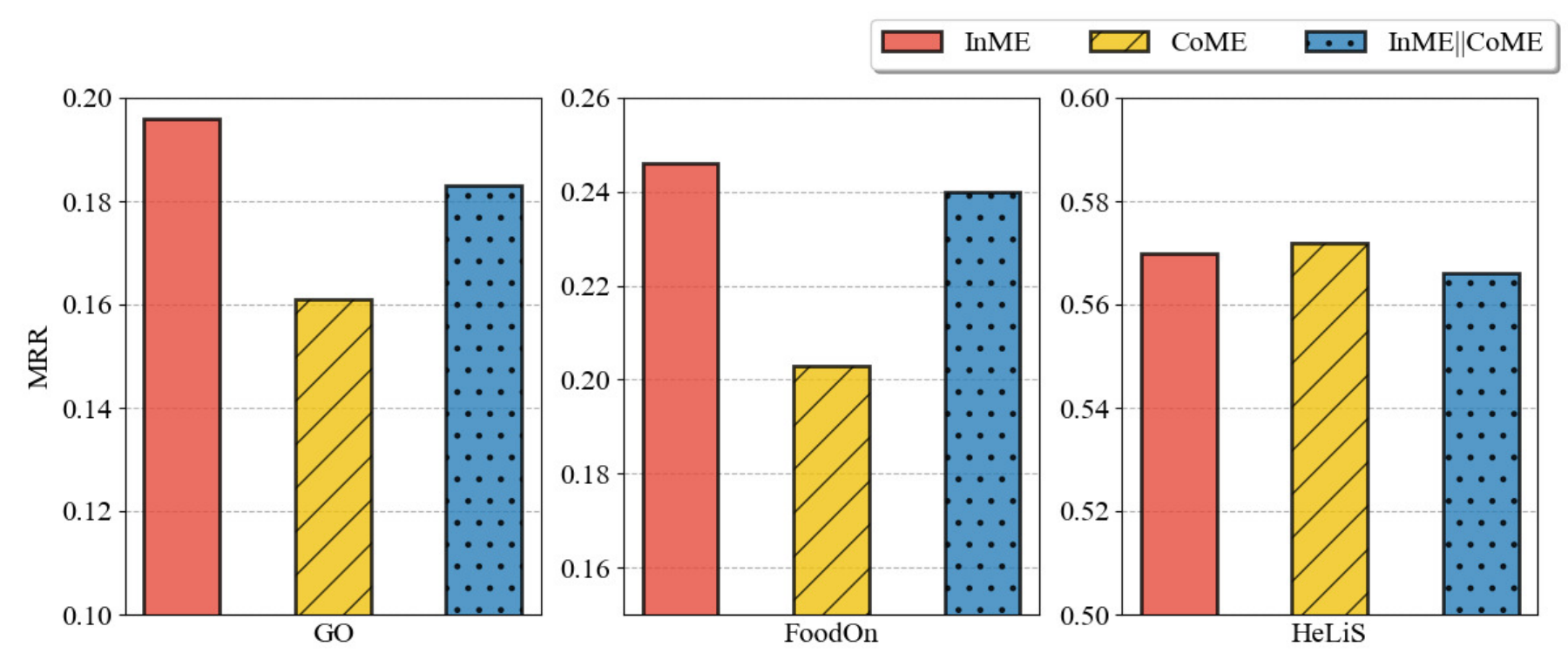}
  \caption{The MRRs of InME, CoME, and InME $||$ CoME.} \label{fig:compare_models}
\end{figure*}
We compare each of our proposed embeddings InME, CoME, and the concatenation of InME and CoME in Figure \ref{fig:compare_models}. For GO and FoodOn, the MRRs of InME are greatly higher than CoME. Also, the MRRs of InME$||$CoME are higher than CoME, but are lower than InME. For HeLiS, all the MRRs are almost the same level because the number of annotation words is insufficient as shown in Table \ref{tab:data}.\par

%
\begin{figure}
  \centering
  \begin{subfigure}{\textwidth}
    \centering
    \includegraphics[width=\textwidth]{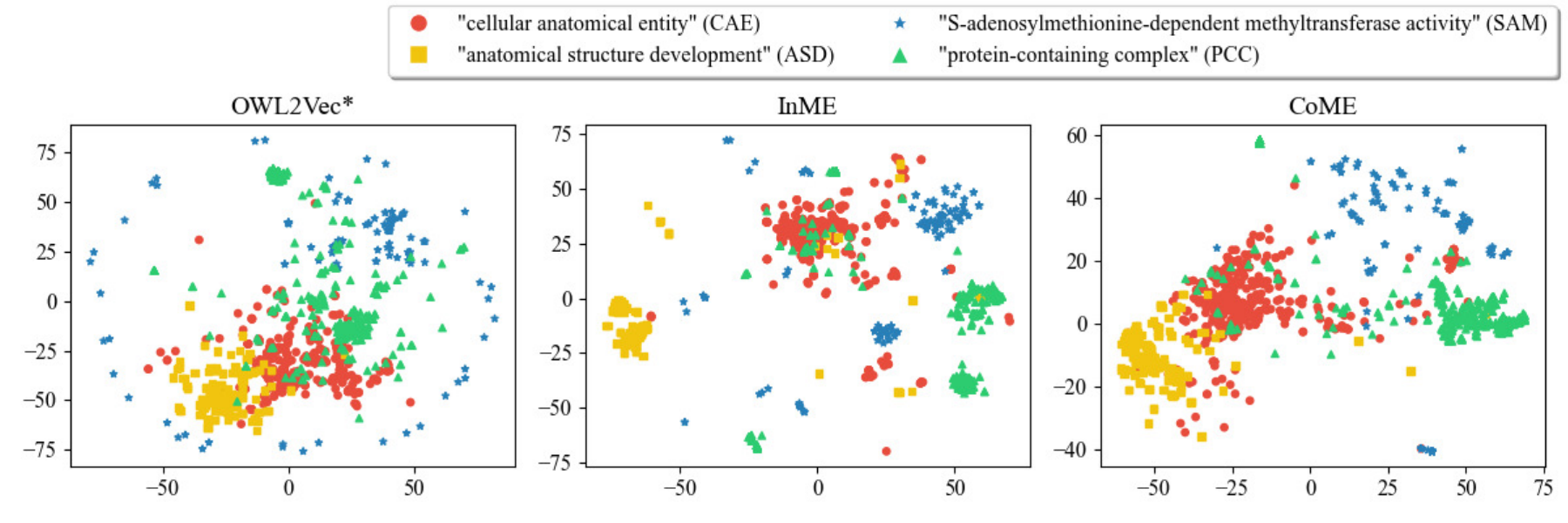}
    \caption{GO}
    \label{fig:scatterplot_go}
  \end{subfigure}

  \begin{subfigure}{\textwidth}
    \centering
    \includegraphics[width=\textwidth]{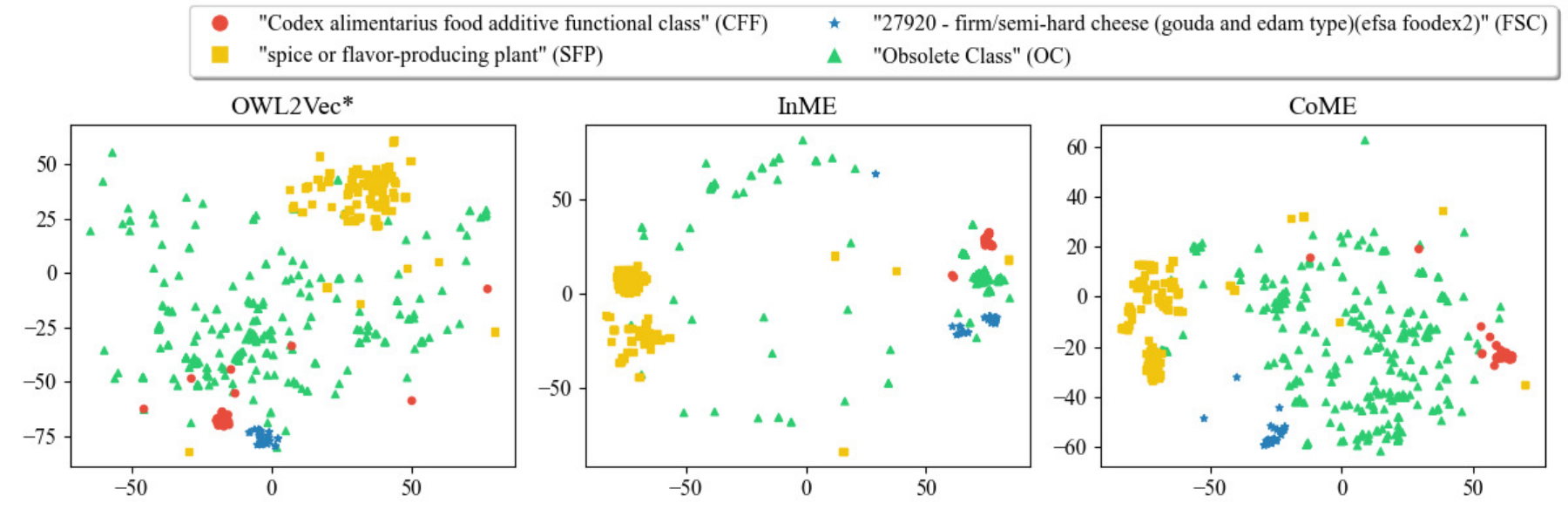}
    \caption{FoodOn}
    \label{fig:scatterplot_foodon}
  \end{subfigure}

  \begin{subfigure}{\textwidth}
    \centering
    \includegraphics[width=\textwidth]{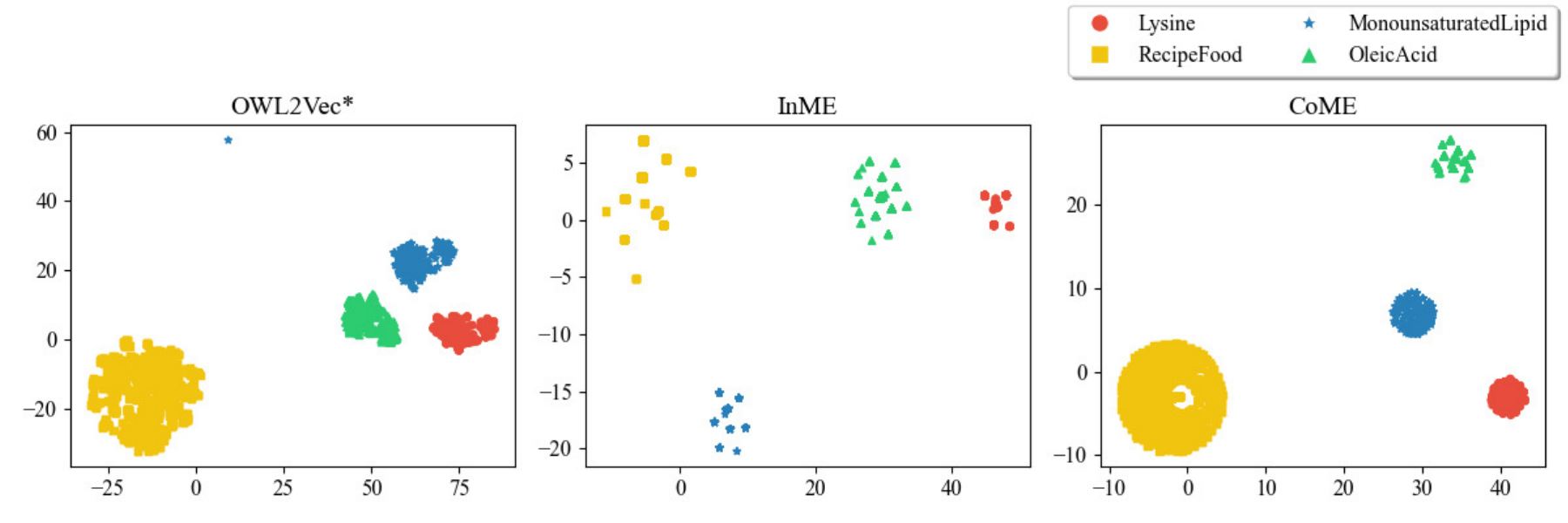}
    \caption{HeLiS}
    \label{fig:scatterplot_helis}
  \end{subfigure}

  \caption{The t-SNE visualizations of the embeddings OWL2Vec*, InME, and CoME on GO, FoodOn, and HeLiS.}
  
  \label{fig:scatterplot}
\end{figure}
Figure \ref{fig:scatterplot} visualizes the embeddings of InME, CoME, and OWL2Vec* on GO, FoodOn, and HeLiS using t-SNE \cite{tsne}. Each point in the figure represents a class (or an individual), with the colors indicating their superclasses. For GO, classes with a common superclass tend to be clustered on the plot in both InME and CoME. In particular, subclasses of SAM (blue points) are circularly distributed in OWL2Vec*, whereas these points are clustered in InME and CoME. For FoodOn, a similar tendency is observed in OWL2Vec* and CoME, where subclasses of OC (green points) are evenly distributed throughout, while the other three subclasses form clusters. On the other hand, we observe that many of the subclasses of OC form a cluster at a single point in InME. These results imply that the global information of InME is more appropriate for capturing the similarity between entities. For HeLiS, the four subclass groups are neatly clustered in InME, CoME, and OWL2Vec*. The embeddings of entities with identical annotation words are converted to the same embeddings through the average of word embeddings by Equation (\ref{eq:word_mean}) or (\ref{eq:word_mean_our}). The well-clustered distributions of HeLiS are largely influenced by the small number of annotations and the averaging embedding transformation method, rather than by the performance of the models.\par
%
\begin{table}[t]
\caption{The results of ablation study of the self-matching training.}
\label{tab:compare_self_matching}
\begin{center}
\resizebox{6cm}{!}{
\begin{tabular}{lc ccc} \toprule
     & & GO & FoodOn & HeLiS \\
    \cmidrule(lr){3-5}
    Model & Self-matching & MRR & MRR & MRR \\ \midrule
    \multirow{2}{*}{InME} 
     & $\checkmark$ & \bf{0.253} & \bf{0.282} & 0.505 \\
     & $\times$ & 0.196 & 0.246 & \bf{0.570} \\ \midrule
    \multirow{2}{*}{CoME}
     & $\checkmark$ & \bf{0.193} & \bf{0.229} & 0.549 \\
     & $\times$ & 0.161 & 0.203 & \bf{0.572} \\ \midrule
\end{tabular}
}
\end{center}
\end{table}
\begin{figure}
    \centering
    \includegraphics[width=8cm]{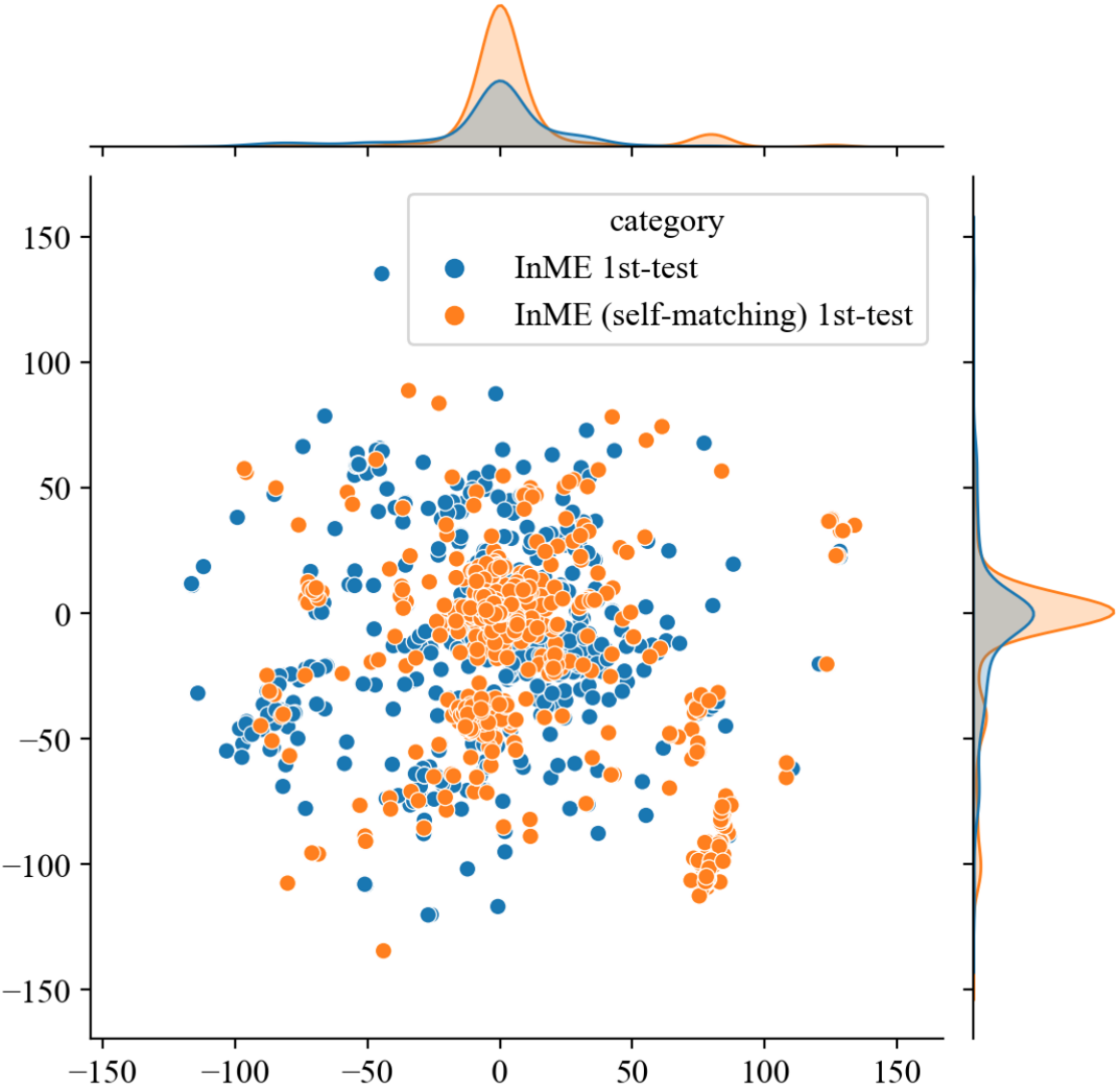}
    \caption{The visualization of first-ranked test data improved by the self-matching training with InME on GO.}
    \label{fig:compare_self_matching}
\end{figure}
Table \ref{tab:compare_self_matching} shows the comparison between the conventional RF training and the self-matching training on the MRRs. The self-matching method improves the MRRs of InME and CoME by 29.0\% and 19.8\% in GO, and 14.6\% and 12.8\% in FoodOn. In Figure \ref{fig:compare_self_matching}, we visualize the embedding InME in GO to analyze how well self-matching improves the prediction performance of similar embeddings. Let $V_{\rm{word}}^{(2)}(e)$ be a two-dimensional entity embedding compressed from $V_{\rm{word}}(e)$ by t-SNE \cite{tsne}. In the figure, we plot $V_{\rm{word}}^{(2)}(D)-V_{\rm{word}}^{(2)}(C)$ of the first-ranked test data $C \sqsubseteq D$, which are improved by the self-matching training with InME on GO. Each plot indicates the distance between the embeddings of a superclass $D$ and subclass $C$. The blue dots represent the first-ranked test data in the conventional RF, and the orange dots represent the test data that newly achieved the first rank by the self-matching training. The clustered orange dots near the origin indicate that the self-matching training improves the prediction performance of similar embeddings. In particular, the similarity representations of InME and CoME are suitable for the self-matching but it does not work well enough for HeLiS with few annotation axioms.\par
%
\begin{table}[t]
\caption{The performance of the self-matching RF classifier (SM-RF) and the conventional RF classifier with InME on restricted test data.}
\label{tab:expr_split_sm}
\begin{adjustwidth}{-1in}{-1in}
\begin{center}
\resizebox{1.0\textwidth}{!}{
\begin{tabular}{l ccc ccc ccc } \toprule
     & \multicolumn{3}{l}{GO} & \multicolumn{3}{l}{FoodOn} & \multicolumn{3}{l}{HeLiS} \\
    \cmidrule(lr){2-4} \cmidrule(lr){5-7} \cmidrule(lr){8-10}
     &  & SM-RF & RF &  & SM-RF & RF &  & SM-RF & RF \\
    Test Data & Num of Data & MRR & MRR & Num of Data & MRR & MRR & Num of Data & MRR & MRR \\ \midrule
    $S_{\rm{test}}$ & 14521 & \bf{0.253} & 0.196 & 5957 & \bf{0.282} & 0.246 & 435 & 0.505 & \bf{0.570} \\
    $S_{\rm{test}}$ on $\mathbb{E}^{+} \!\times\! \mathbb{E}^{+}$ & 9702 & \bf{0.270} & 0.204 & 1543 & \bf{0.205} & 0.192 & 0   & -          & -          \\
    $S_{\rm{test}}$ on $\mathbb{E}^{I} \!\times\! \mathbb{E}^{+}$ & 4456 & \bf{0.211} & 0.189 & 4235 & \bf{0.313} & 0.274 & 429 & 0.511      & \bf{0.577} \\
    $S_{\rm{test}}$ on $\mathbb{E}^{+} \!\times\! \mathbb{E}^{I}$ & 203  & \bf{0.331} & 0.061 & 59   & \bf{0.135} & 0.028 & 0   & -          & -          \\
    $S_{\rm{test}}$ on $\mathbb{E}^{I} \!\times\! \mathbb{E}^{I}$ & 160  & \bf{0.218} & 0.035 & 120  & \bf{0.243} & 0.017 & 6   & \bf{0.023} & 0.009      \\
    \bottomrule
\end{tabular}
}
\end{center}
\end{adjustwidth}
\end{table}
We analyze the concept subsumption prediction for isolated entities using the self-matching RF classifier (SM-RF) and the conventional RF classifier. Table \ref{tab:expr_split_sm} shows the comparison of the two methods on restricted test data, where $\mathbb{E}^{+}$ is the set of entities in $S_{\rm{train}}^{+}$ and $\mathbb{E}^{I}$ is the set of isolated entities as $\mathbb{E} \setminus \mathbb{E}^{+}$. The SM-RF achieves the MRRs of 0.331 and 0.135 on $\mathbb{E}^{+} \!\times\! \mathbb{E}^{I}$, and 0.218 and 0.243 on $\mathbb{E}^{I} \!\times\! \mathbb{E}^{I}$ for GO and FoodOn but the conventional RF fails to predict some isolated entities on the restricted test data.

\section{Conclusion \label{sec:conclusion}}
In this paper, we proposed the self-matching training method with the two embeddings InME and CoME for concept subsumption prediction. Our method overcomes the problem that conventional Random Forest training with OWL2Vec* fails to predict the embeddings of isolated superclasses from the embeddings of similar subclasses. The important point is that the self-matching training can learn all the entities including isolated entities but the conventional training learns only the entities used in training data. In ontology embeddings, we showed that the global and local information of annotation axioms in InME and CoME are more effective than logical axioms for the concept subsumption prediction. Our evaluation experiments demonstrated that the self-matching training method with InME and CoME $||$ OWL2Vec* outperforms the existing models on benchmark ontologies. \par
In future work, there is further research to leverage the logical axioms in our method to improve the prediction performance and extend to description logic reasoning tasks, such as predicting more complex classes (e.g., $C \sqsubseteq \exists R.D$). Additionally, we plan to apply the self-matching training method to inductive prediction tasks for various relationships and assertions in ontologies and knowledge graphs.

\appendix
\section{Parameter Settings \label{sec:append}}
Table \ref{tab:append} displays the hyper-parameters selected by the performance of the validation data in our experiments, as shown in Table \ref{tab:result_compare}.\par

\begin{table}[htb]
    \caption{The best hyper-parameter settings of each model.}
    \label{tab:append}
    \begin{adjustwidth}{-1in}{-1in}
    \begin{center}
    \resizebox{1.0\textwidth}{!}{ 
    \begin{tabular}{llcccccc} \toprule
        \multicolumn{2}{l}{} & \multicolumn{2}{l}{GO} & \multicolumn{2}{l}{FoodOn} & \multicolumn{2}{l}{HeLiS} \\
        \cmidrule(lr){3-4} \cmidrule(lr){5-6} \cmidrule(lr){7-8}
        Model & Concatenation & Dim & Self-matching
        & Dim & Self-matching
        & Dim & Self-matching \\ \midrule
        \multirow{3}{*}{InME} & - & 100 & $\checkmark$ 
            & 50 & $\checkmark$ 
            & 100 & $\times$ \\ 
         & OWL2Vec* & 100 & $\checkmark$ 
            & 50 & $\checkmark$ 
            & 100 & $\times$ \\ 
         & Word2Vec & 100 & $\checkmark$ 
            & 50 & $\checkmark$ 
            & 200 & $\times$ \\ \midrule 
        \multirow{4}{*}{CoME} & - & 100 & $\checkmark$ 
            & 50 & $\checkmark$ 
            & 200 & $\checkmark$ \\ 
         & OWL2Vec* & 100 & $\checkmark$ 
            & 50 & $\checkmark$ 
            & 100 & $\times$ \\ 
         & Word2Vec & 50 & $\checkmark$ 
            & 50 & $\checkmark$ 
            & 200 & $\checkmark$ \\ 
         & InME & 100 & $\checkmark$ 
            & 50 & $\checkmark$ 
            & 50 & $\times$ \\ \bottomrule 
    \end{tabular}
    }
    \end{center}
    \end{adjustwidth} 
\end{table}

\bibliographystyle{junsrt} 
\bibliography{reference}

\end{document}